\documentclass[letterpaper, 10 pt, conference]{ieeeconf}  

\IEEEoverridecommandlockouts                              

\usepackage{graphics} 
\usepackage{graphicx}
\usepackage{algorithm}
\usepackage{algorithmic}
\usepackage{amsmath} 
\usepackage{multirow}
\usepackage{hyperref}

\title{\LARGE \bf
ManipGPT: Is Affordance Segmentation by Large Vision Models Enough for Articulated Object Manipulation?
}

\author{Taewhan Kim$^{1,2}$, Hojin Bae$^{1}$, Zeming Li$^{1}$, Xiaoqi Li$^{1,2}$, Iaroslav Ponomarenko$^{1,2}$, Ruihai Wu$^{1}$, Hao Dong$^{1,2}$%
\thanks{$^{1}$CFCS, School of Computer Science,
Peking University $^{2}$PKU-Agibot Lab; Correspondence to Hao Dong {\tt\small hao.dong@pku.edu.cn}}
}

\begin{document}

\maketitle
\thispagestyle{empty}
\pagestyle{empty}

\begin{abstract}

Visual actionable affordance has emerged as a transformative approach in robotics, focusing on perceiving interaction areas prior to manipulation. Traditional methods rely on pixel sampling to identify successful interaction samples or processing pointclouds for affordance mapping. However, these approaches are computationally intensive and struggle to adapt to diverse and dynamic environments. This paper introduces ManipGPT, a framework designed to predict optimal interaction areas for articulated objects using a large pre-trained vision transformer (ViT). We create a dataset of 9.9k simulated and real images to bridge the visual sim-to-real gap and enhance real-world applicability. By fine-tuning the vision transformer on this small dataset, we significantly improve part-level affordance segmentation, adapting the model’s in-context segmentation capabilities to robot manipulation scenarios. This enables effective manipulation across simulated and real-world environments by generating part-level affordance masks, paired with an impedance adaptation policy, sufficiently eliminating the need for complex datasets or perception systems. Our project page is available at: \href{https://lxkim814.github.io/ManipGPT_website/}{\texttt{\detokenize{https://lxkim814.github.io/ManipGPT_website/}}}

\end{abstract}

\section{INTRODUCTION}
Robotic manipulation is crucial for enabling robots to perform in wide variety of tasks autonomously. Despite its importance, it remains a complex and challenging problem largely due to the variability in objects, mechanisms, and environmental constraints.

A promising approach to address this challenge is visual actionable affordance, which reframes robotic manipulation as a perception task. This method uses visual cues to predict optimal interaction points, which streamlines manipulation tasks. Traditional affordance map generation~\cite{mo2021where2act, wu2021vat,xu2022universal} involves sampling pixels on an object, interacting at each point, and recording scores to construct a map. This process is repetitive, computationally intensive, and further burdened by reliance on point cloud data. Additionally, the interaction data collected from simulated objects have physical and visual discrepancies from real objects, creating a significant sim-to-real gap.
\begin{figure}[h]
    \begin{center}
        \includegraphics[width=\linewidth]{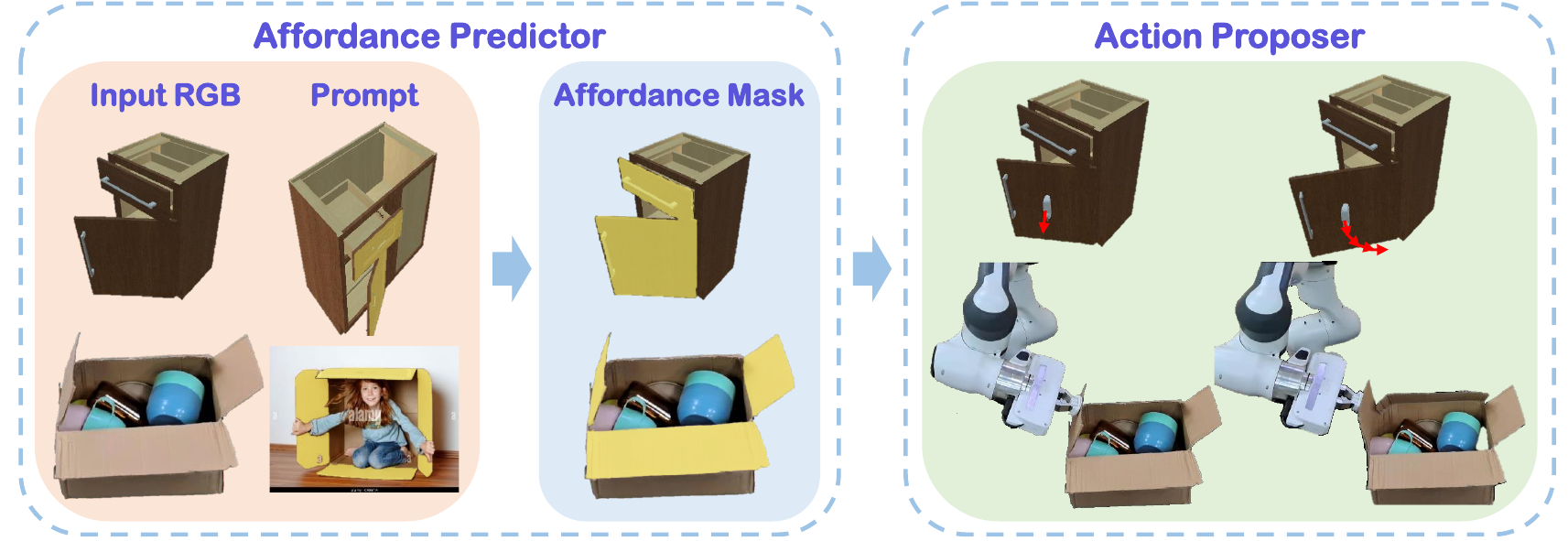}
    \end{center}
    \caption{ManipGPT processes an RGB image with a visual prompt to generate an affordance mask, which determines the contact point and manipulation direction.}
    \label{fig:teaser}
\end{figure}

Recent advancements leverage large-scale datasets and large language models (LLMs) to enhance task execution~\cite{huang2023voxposer,li2024manipllm}. However, these methods often result in high computational costs and increased complexity, limiting their practicality in resource-constrained environments. Additionally, a gap between language and vision modalities hinders the seamless integration of visual and language-based commands, affecting the execution of complex tasks.

To overcome these limitations, we present ManipGPT, a minimalist approach that leverages a fine-tuned large vision model in context to predict affordance masks for a wide range of indoor objects, significantly simplifying robotic manipulation. Our streamlined approach, using a single RGB image and a pair of prompt images to directly generate affordance masks, eliminates the need for iterative sampling techniques and enables one-shot manipulation without requiring robot-specific training. Additionally, we present a synthetic dataset of 9.9k images from both simulated and realistic environments, demonstrating that even minimal datasets with basic annotations can enhance robotic performance effectively.

In this work, we define an affordance representation as any manipulatable geometry or region of an object that enables interaction to achieve a desired motion. By segmenting such regions, our approach generalizes across different object instances and enables visual affordance prediction without requiring explicit action annotations.

Our system integrates affordance masks with a post-processing algorithm to ensure safe and efficient manipulation, including an optional impedance control mechanism for improved physical adaptability. Experimental results show that our framework predicts interaction points with human-like intuition, generating actionable affordance maps that enable robots to identify contact areas and motion directions.

\section{RELATED WORKS}

\subsection{Articulated Object Manipulation}

Articulated object manipulation poses significant challenges due to geometric variability and the interactive nature of object parts. Existing methods largely fall into two categories: affordance-based policies and articulation estimation.

Affordance-based policies identify actionable regions on objects by generating affordance maps that indicate the likelihood of moving object parts~\cite{mo2021where2act, wu2021vat, xu2022universal, eisner2022flowbot3d}. While these methods offer generalization to novel objects, they typically involve a dataset collection stage where a robot engages in trial-and-error interactions by sampling points in a point cloud or pixels in an image, aiming to identify the most actionable regions. This process is time-consuming and computationally intensive, especially in a 3D simulation environment. Moreover, the reliance on simulations introduces a sim-to-real gap, meaning the affordance map collected from simulated interactions may not accurately reflect real-world scenarios.

Articulation estimation methods, on the other hand, predict 6D joint parameters directly from visual inputs such as a sequence of images~\cite{jain2021screwnet, jiang2022ditto}, RGB-D images~\cite{heppert2022category,zeng2021visual}, or point clouds~\cite{fu2024capt,li2020category}. By segmenting images or point clouds and estimating part poses, these methods enable robust manipulation trajectories.

In contrast, our approach simplifies the process by using a single RGB image and prompt inputs to generate an affordance mask. This eliminates the need for pointclouds or iterative sampling for affordance learning and achieves effective manipulation with minimal computational overhead.

\subsection{Large Models and Applications in Robotics}


Recent advancements in large vision models~\cite{kirillov2023segment,wang2023seggpt,ke2024segment} have advanced semantic and instance segmentation. These models allow robots to operate in diverse environments with minimal real-world training. By leveraging pre-trained models, robots can quickly generalize to new scenarios and improve operational efficiency. In robotics, such models are especially valuable for tasks that require fine-grained visual understanding, such as point cloud and 2D image segmentation~\cite{zhang2024sam,huang2024manipvqa}.


LLMs like GPT~\cite{brown2020language} and LLaMA~\cite{touvron2023llama} have enhanced robotics by enabling natural language understanding and high-level reasoning. When integrated with Visual Language Models (VLMs), they strengthen robots to interpret language commands and execute tasks with contextual awareness and adaptability~\cite{huang2023voxposer,li2024manipllm}.

While large vision models have widely been used as supplementary tools for pre-processing or post-processing, with LLMs serving as the backbone, our approach diverges by implementing SegGPT~\cite{wang2023seggpt} as the primary backbone model. Unlike resource-intensive LLMs, SegGPT's smaller 370 million parameter architecture offers vivid segmentation and flexible visual prompting, making it an efficient application.

\subsection{2D Affordance Datasets}
Existing 2D affordance datasets primarily focus on small, hand-held objects with well-defined interaction points, such as kitchen utensils and tools~\cite{guo2023handal,luo2022learning,nguyen2017object}. While these datasets provide affordance annotations, they are limited to single-part objects and do not capture affordance regions on articulated objects with multiple manipulatable components. Articulated objects, such as drawers and refrigerators, require affordance segmentation that identifies distinct parts (e.g., doors, lids, drawers). This highlights the need for datasets that support part-level affordance segmentation in articulated objects.

\section{Method}

\begin{figure*}[!h]
    \begin{center}
        \includegraphics[width=1\textwidth]{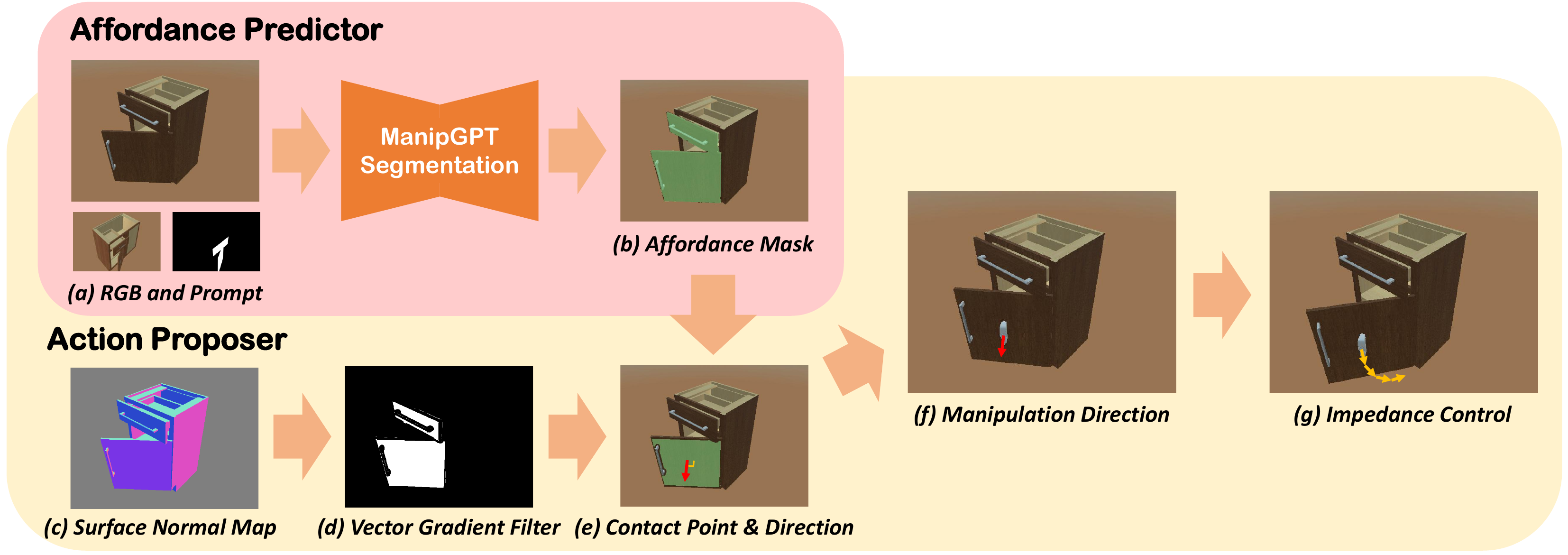}
    \end{center}
    \caption{The pipeline is divided into two main modules: Affordance Predictor and Action Proposer. (a) The RGB image and category-specific prompt are input to generate an affordance mask. (b) The affordance mask highlights actionable parts of the object. (c) A surface normal map is used to understand the object's surface orientation. (d) A vector gradient filter refines the normal map by filtering out non-ideal areas. (e) The optimal contact point and manipulation direction are identified. (f) The manipulation direction is set for the gripper's action. (g) An impedance control algorithm guides the robot’s manipulation based on physical feedback.}
    \label{fig:pipeline}
\end{figure*}

Our framework is divided into two key modules: the Affordance Predictor and the Action Proposer. The model-based Affordance Predictor generates an affordance mask from an RGB input, while the algorithm-based Action Proposer identifies the optimal grasping spot, initial manipulation direction, and subsequent action directions.

\subsection{Affordance Predictor}

Our affordance predictor generates part-level affordance masks for articulated objects using a one-shot segmentation approach. Given a query RGB image and a category-specific visual prompt, the model encodes both inputs jointly to compare local features and identify regions in the query image that correspond to the affordance shown in the prompt. As shown in Figure~\ref{fig:pipeline}(a)(b), the model extracts visual features from the storage furniture and aligns them with the door and drawer masks observed in the prompt. This in-context learning mechanism enables the system to transfer affordance knowledge across similar object instances.

This module identifies manipulatable geometry rather than predefined action points. The affordance mask highlights regions that are expected to support interaction based on their prompts. This differs from traditional affordance prediction models that explicitly assign labels such as ``push'' and ``pull.'' Instead, our approach extracts geometric cues from the scene, allowing downstream components to dynamically infer appropriate manipulation strategies.

In  real robot experiments, we refine affordance masks using SAM~\cite{kirillov2023segment} for multi-part segmentation, while in simulation we use built-in segmentation to select a single part. To maintain consistency, we select the largest segmented region, ensuring robust interaction with the object while avoiding physically constrained areas.

\subsection{Action Proposer}

After identifying the actionable region, the next step is to translate these visual cues into physical manipulation. We achieve this by extracting local orientation cues from the segmented region using surface normal vectors. This geometric information is essential for estimating the direction of interaction and guiding the end-effector pose. In articulated objects, motion typically occurs along the normal of the contact area, making it a strong cue for effective manipulation, as demonstrated in prior work leveraging such information for affordance data collection~\cite{mo2021where2act,wu2021vat}.

However, when contact points lie on irregular or curved surfaces, direct use of these vectors can be unreliable. To address this, we apply gradient filtering on the normal map to remove noisy areas and extract a stable contact direction, as detailed in Algorithm~\ref{alg:algorithm1}. For example, in the case of storage furniture with a cylindrical handle at the center (Figure~\ref{fig:pipeline}(d)), the raw orientation data may be suboptimal. Filtering allows us to locate a more reliable region for defining the end-effector pose, interaction point, and manipulation direction.

The remaining manipulation steps are executed using the active impedance control algorithm from~\cite{li2024manipllm}, which enables precise motion control, particularly for interacting with revolute joints. By following a step-by-step policy, the robot adjusts its grip and motion in real time based on physical feedback. This allows it to adapt to various object shapes, joint types, and constraints without requiring further reasoning or explicit trajectory prediction.

\begin{algorithm}[tb]
    \caption{Finding contact point and direction for manipulation}
    \label{alg:algorithm1}
    \textbf{Input}: Normal map $N_{map}$, part mask $M_{part}$\\
    \textbf{Output}: Contact point $(x, y)$, manipulation direction $\mathbf{n}$
    \begin{algorithmic}[1] 
        \STATE Apply gaussian blur to smooth the normal map $N_{map}$.
        \STATE Split the normal map $N_{map}$ into x, y, and z components $(N_x, N_y, N_z)$ and compute gradients $\nabla_x$, $\nabla_y$, and $\nabla_z$.
        \STATE Calculate the gradient magnitude $G_{mag} = \sqrt{\nabla_x^2 + \nabla_y^2 + \nabla_z^2}$.
        \STATE Generate edge mask $M_{edge}$ by thresholding $G_{mag}$ with a predefined filter value.
        \STATE Multiply inverted $M_{edge}$ by $M_{part}$ to create $M_{invedge}$, a mask for flat surfaces.
        \STATE Apply $M_{invedge}$ to the original normal map $N_{map}$ to obtain the masked normal map $N_{masked}$.
        \STATE Compute the centroid $(c_x, c_y)$ of $M_{part}$.
        \IF {$(c_x, c_y) \in N_{masked}$ \textbf{and} $N_{masked}(c_x, c_y) \neq [0, 0, 0]$}
        \STATE Set the contact point $(x, y) = (c_x, c_y)$.
        \STATE Set the manipulation direction $\mathbf{n}$ to the normal vector at $(c_x, c_y)$.
        \ELSE
        \STATE Set the manipulation direction $\mathbf{n}$ to the most frequent non-zero normal vector in $N_{masked}$.
        \IF {$N_{masked}(c_x, c_y) = [0, 0, 0]$}
        \STATE Define a centered bounding box $B$ with one-third dimensions of $M_{part}$ around $(c_x, c_y)$.
        \IF {$\forall (x, y) \in B, \ N_{masked}(x, y) \neq [0, 0, 0]$}
        \STATE Set $(x, y)$ to a random non-zero pixel in $B$.
        \ELSE
        \STATE Set $(x, y)$ to a random non-zero pixel in $M_{part}$.
        \ENDIF
        \ENDIF
        \ENDIF
        \STATE \textbf{return} $(x, y)$, $\mathbf{n}$
    \end{algorithmic}
\end{algorithm}

\section{Dataset}

\begin{figure}[h]
    \centering
    \includegraphics[width=0.4\textwidth]{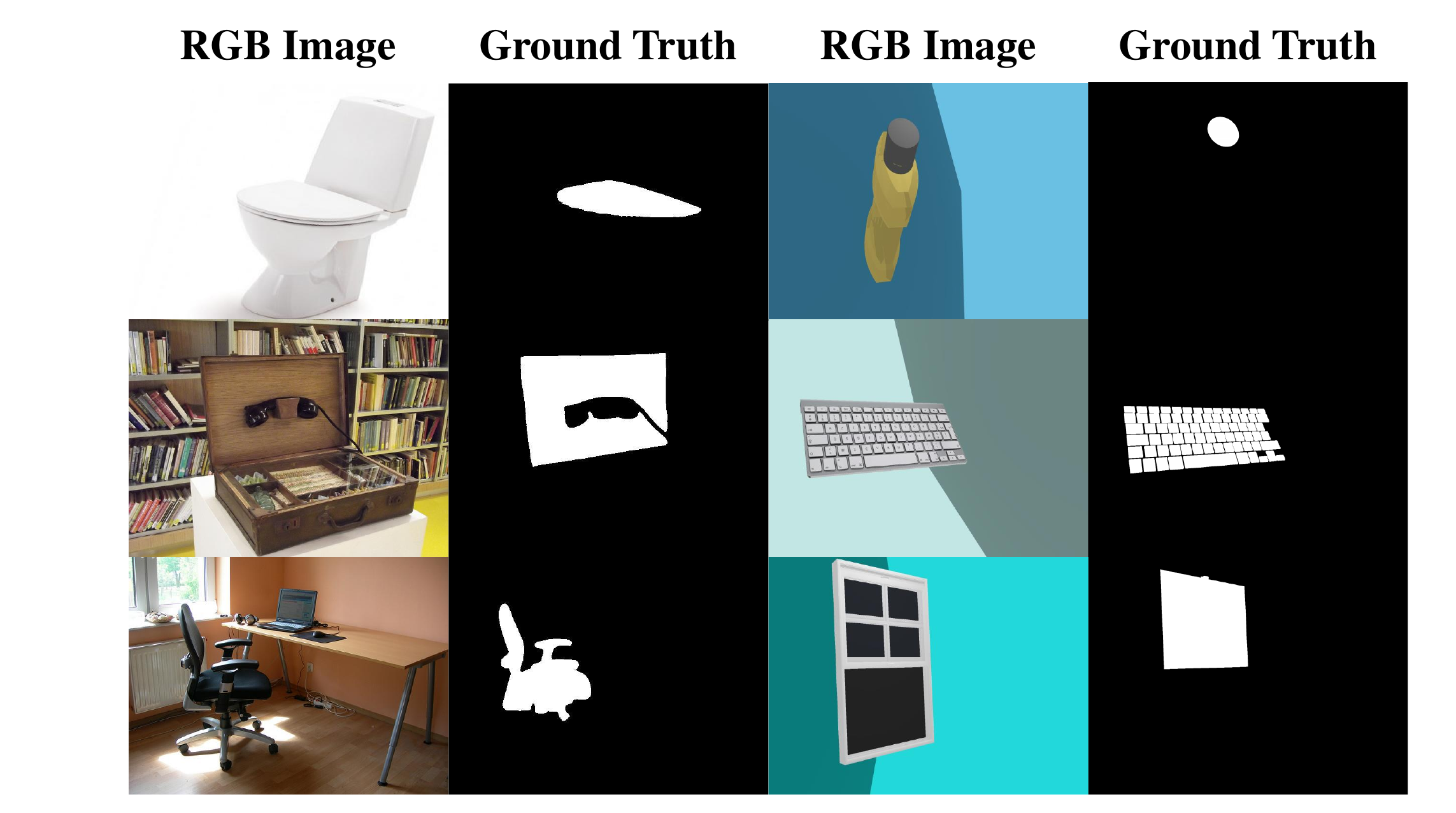}
    \caption{Sample RGB images and ground truth affordance masks: real-world examples (left) and simulated images (right). White regions indicate manipulable affordance areas.}
    \label{fig:dataset}
\end{figure}

We construct a dataset tailored for affordance detection in robotic manipulation, comprising 13,350 images across 30 object categories—18 for training and 12 for testing. The dataset integrates simulated and real-world images to ensure robustness and adaptability, as shown in Figure \ref{fig:dataset}.

\textbf{Training Data:} The training set includes 9,900 images: 9,000 simulated images generated in the SAPIEN environment~\cite{xiang2020sapien} and 900 real-world images. Simulated data spans 18 categories, with 5 objects per category and 100 articulation steps captured from random angles (5 × 100 × 18). Real-world images, sourced from ImageNet~\cite{deng2009imagenet}, COCO~\cite{lin2014microsoft}, OpenImagesV7~\cite{kuznetsova2020open}, MyNursingHome~\cite{ismail2020mynursinghome}, Pixabay~\cite{Pixabay}, and our own captures, include 50 images per category.

\textbf{Testing Data:} The test set comprises 3,450 images: 3,000 simulated images sampled from random viewpoints and articulations of novel instances and 450 real-world images. Real-world images are annotated using CVAT.ai~\cite{CVAT}, while simulated annotations are automatically derived from actionable link IDs.

Annotation criteria ensure masks encompass the manipulable parts, focusing where a robot can attach and interact. Minor overlaps with non-relevant surfaces are allowed if the primary focus remains on the manipulable part.

By combining controlled simulated environments with diverse real-world data, this dataset provides a foundation for advancing affordance detection in manipulation.

\section{Experiment}

\subsection{Training Details} 
 The fine-tuning process is crucial for adapting our backbone model from a general segmentation framework to a specialist in part-level affordance detection, enabling a robot arm to approach and execute precise manipulations. We train the model for 300 epochs with a learning rate of 1e-5, keeping all other hyper-parameters identical to those of the backbone model, including the L1 loss function. The training is conducted on a single A800 80G GPU.

For in-context tuning, we create specific prompts for 30 categories outside our dataset. These prompts are designed to help the model better identify affordance regions in a variety of objects. For each category, we use one real image and one simulated image as prompts to test the model on both real and simulated data. The prompt images come from the same domain as the train/test data but contain different object instances.

\begin{table}[h!]
  \begin{center}
  \caption{Segmentation result on simulated and real images. ``*'' indicates test result on real images only.}
    \label{tab:compasion_on_sapien&real_img}
    \begin{tabular}{l|c|c|c|c} 
    \textbf{Method} & \textbf{mIoU} & \textbf{F}\textsubscript{1}\textbf{score} & \textbf{mIoU*} & \textbf{F}\textsubscript{1}\textbf{score*}\\
      \hline
      ViT  & 34.1 & 37.2 & 14.8 & 26.0\\
      Bayesian  & 47.0 & 56.9 & 55.5 & 65.0\\
      SegGPT  & 34.9 & 56.0 & 36.2 & 45.1\\
      LISA & 46.3 & 56.2 & 40.1 & 49.1\\
      Ours & \textbf{66.5} & \textbf{74.5} & \textbf{73.4} & \textbf{80.3}\\
    \end{tabular}
  \end{center}
\end{table}

\begin{figure*}[h]
\centering
\includegraphics[width=1.0\textwidth]{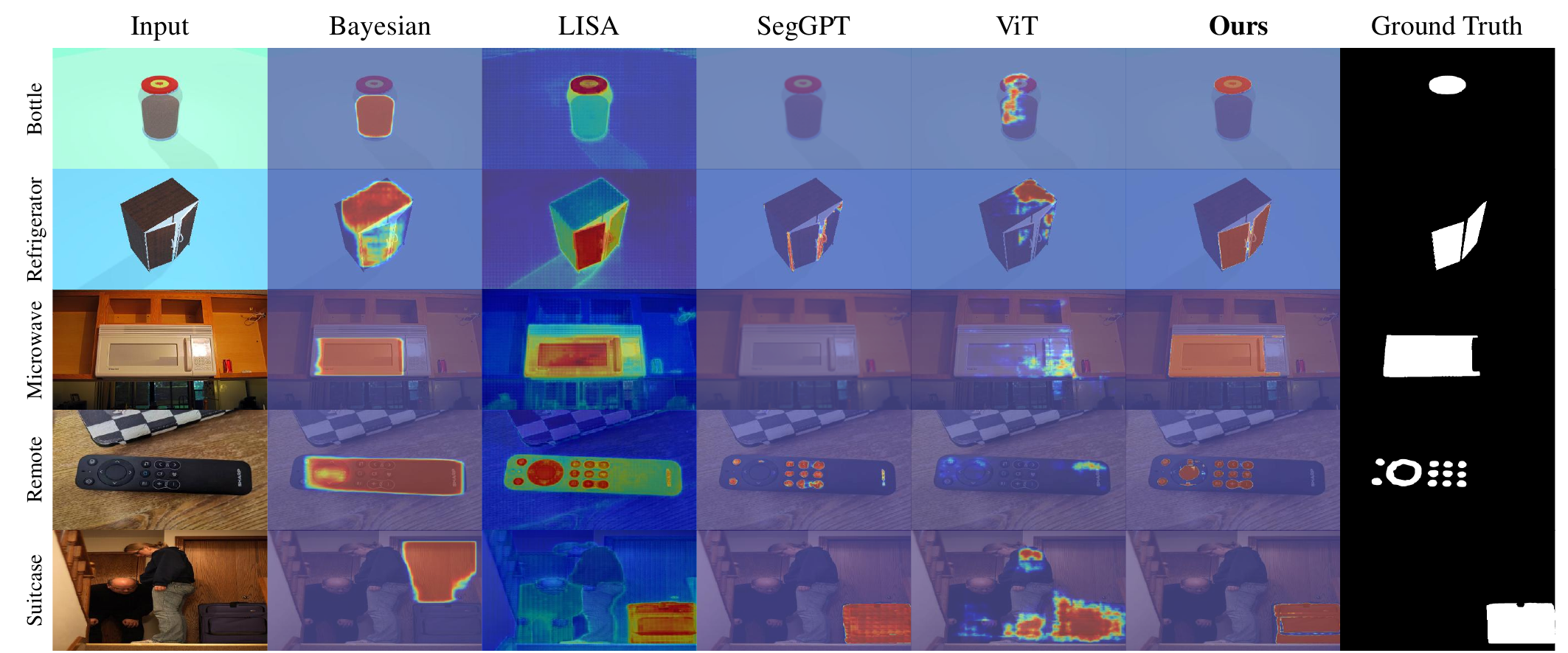}
\caption{Thermal map comparison of ManipGPT and other models. LISA struggles to mask out all actionable parts for certain categories, such as doors. Other models are lack of generalization ability.}
\label{fig}
\end{figure*}

\subsection{Affordance Segmentation Experiment and Baselines} 
We evaluate our model's performance against several segmentation baselines, summarized as follows:

\begin{itemize}
    \item \textbf{Vision Transformer}: Inspired by~\cite{dosovitskiy2020image}, our custom ViT model divides images into 16×16 patches, encoding each patch into a 768-dimensional vector. These vectors are processed through 12 attention blocks, each with 8 attention heads and a multi-layer perceptron. The model is trained for 500 epochs using a cross-entropy loss function, with an initial learning rate of 4e-5 and a dropout rate of 0.2.
    \item \textbf{Bayesian}~\cite{mur2023bayesian}: Using ResNet-50 blocks, this model identifies multiple affordance masks within a single image. We adapt it to detect part-level affordance mask, training it on our dataset without modifying its original hyperparameters.
    \item \textbf{LISA}~\cite{lai2024lisa}: A multi-modal, language-prompted segmentation model with advanced reasoning capabilities. LISA can segment various elements using natural language prompts. Since it requires a language dataset and is already pre-trained for part segmentation, we test LISA without additional fine-tuning to assess the performance of language-prompted segmentation against image-prompted approaches.
    \item \textbf{SegGPT}~\cite{wang2023seggpt}: Serving as our backbone model, SegGPT is tested to benchmark its affordance segmentation performance. This comparison highlights the extent of improvement achieved by our fine-tuning approach.
\end{itemize}

\subsection{Baseline Comparison and Analysis}
The vanilla ViT, due to its small scale, show rapid loss reduction during training, but it struggles to generalize to unseen categories, particularly in real-world images. This limitation makes it less suitable for real-world applications.

Similarly, the limited diversity of training categories constrains the Bayesian model’s segmentation capability;  although it performs well on its own test set, it lacks generalization. Training the Bayesian model from scratch will also require substantial data collection and computational resources, leading to poorer performance compared to ManipGPT when using the same training set.


Despite the use of carefully designed language prompts to guide segmentation, LISA's performance falls short compared to both ManipGPT and SegGPT. For instance, when given the prompt ``segment all flaps or wings of the box,'' LISA often misinterprets the instruction, segmenting the entire box instead of the specified parts. This highlights a limitation of language prompts in providing precise guidance for vision tasks, where ambiguity in language can lead to errors. In contrast, image prompts inherently offer a clearer and more aligned context for segmentation, avoiding the challenges posed by modality switching from language to vision.

ManipGPT outperforms all baselines, achieving higher mean Intersection over Union (mIoU) and F1-scores on both simulated and real images (Table~\ref{tab:compasion_on_sapien&real_img}). By fine-tuning on a small dataset and leveraging SegGPT’s pretrained weights, it delivers accurate and generalizable affordance segmentation in real-world settings.

\begin{table*}[tb]
		
	\begin{center}
	\caption{Comparison of our method against baseline models. `Ours (random)' refers to the random point policy, while `Ours (long)' refers to multi-step manipulation. All other methods are evaluated using one-step manipulation.} 
 \label{tab:pull}	
	\small
	    \setlength{\tabcolsep}{1.5mm}{
		\begin{tabular}{c| c c c c c c c c c c c c c c c c}
  \hline
 \multirow{2}{*}{\textbf{}}&\multirow{2}{*}{\textbf{}} &\multicolumn{15}{c}{\textbf {Novel Instances in Train Categories}}\\
 Method
    &\includegraphics[height=0.035\linewidth]{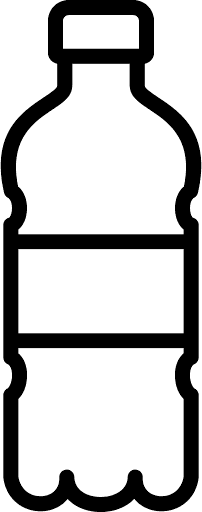}
        &\includegraphics[height=0.035\linewidth]{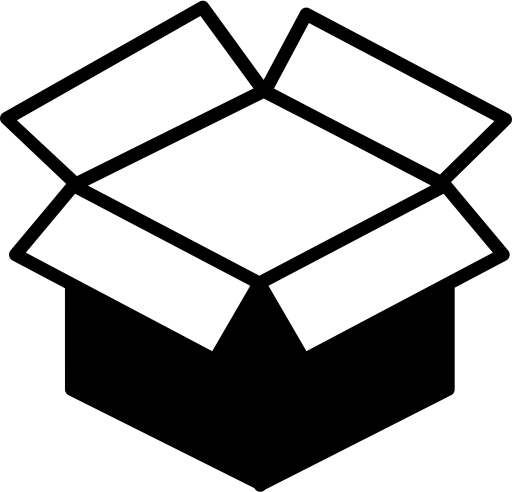}
        &\includegraphics[height=0.035\linewidth]{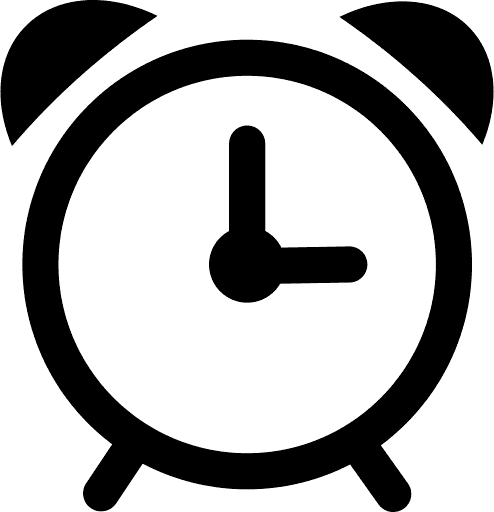}
        &\includegraphics[height=0.035\linewidth]{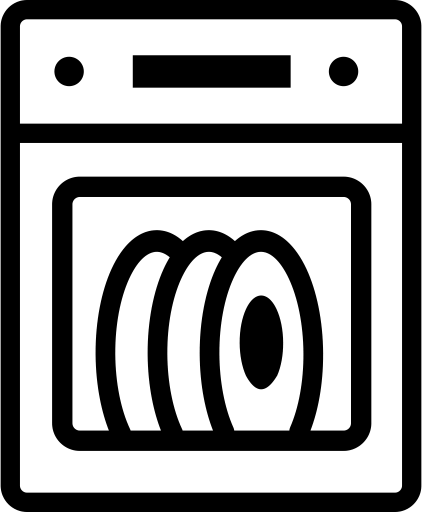}
        &\includegraphics[height=0.035\linewidth]{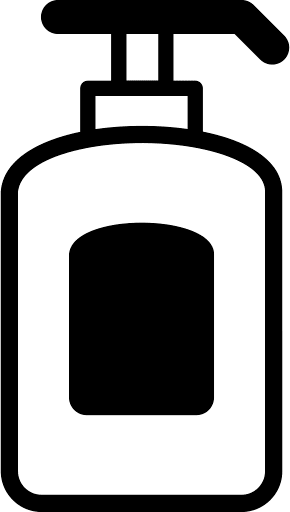}
        &\includegraphics[height=0.035\linewidth]{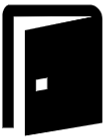}
        &\includegraphics[height=0.035\linewidth]{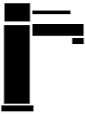}
        &\includegraphics[height=0.02\linewidth]{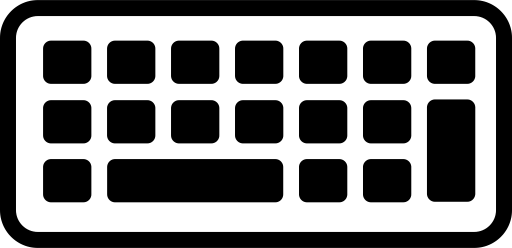}
        &\includegraphics[height=0.035\linewidth]{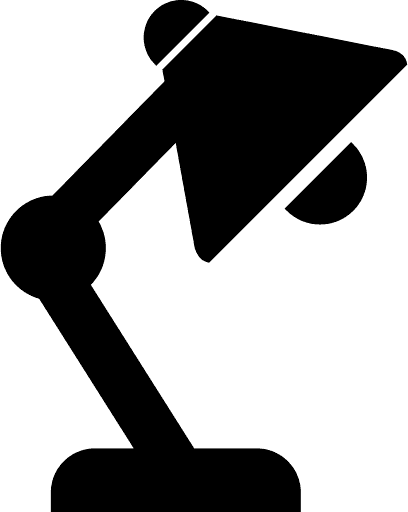}
        &\includegraphics[height=0.035\linewidth]{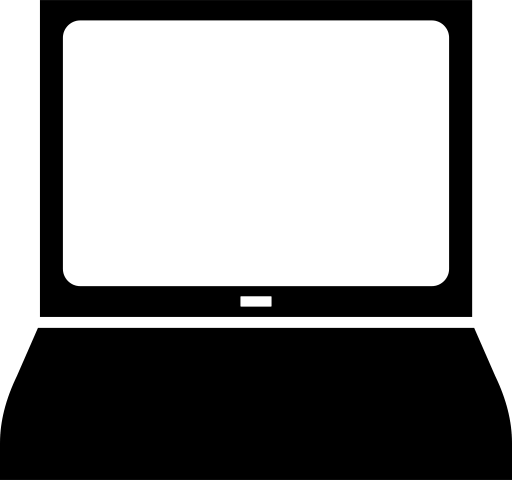}
        &\includegraphics[height=0.03\linewidth]{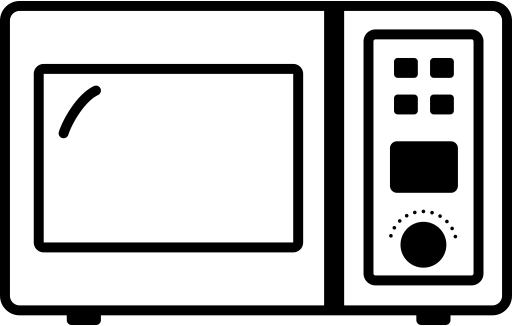}
        &\includegraphics[height=0.035\linewidth]{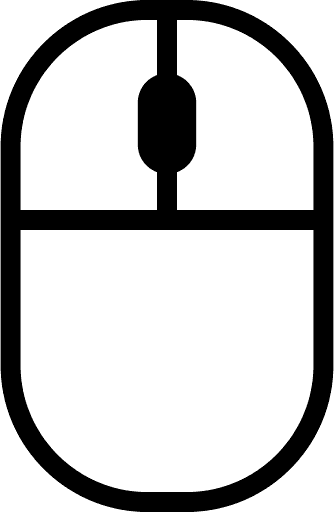}
        &\includegraphics[height=0.035\linewidth]{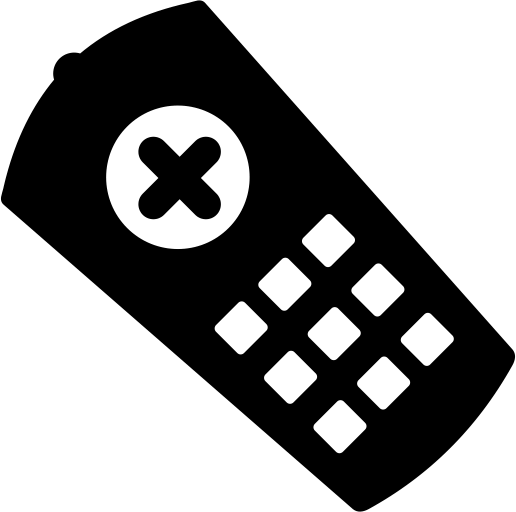}
        &\includegraphics[height=0.035\linewidth]{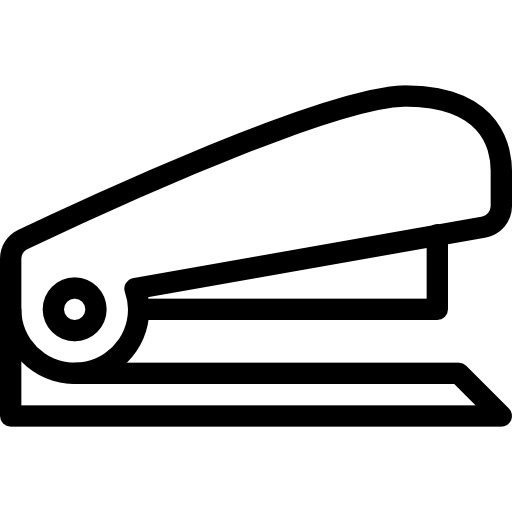}
        &\includegraphics[height=0.035\linewidth]{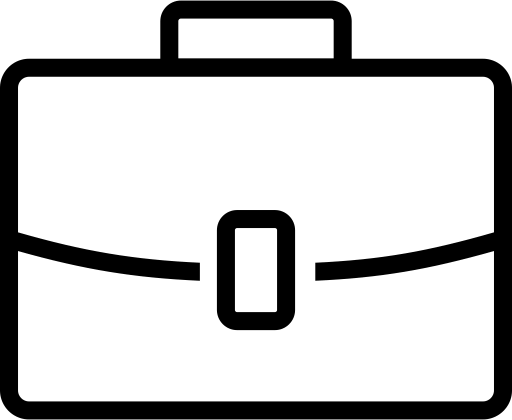}
        &\includegraphics[height=0.035\linewidth]{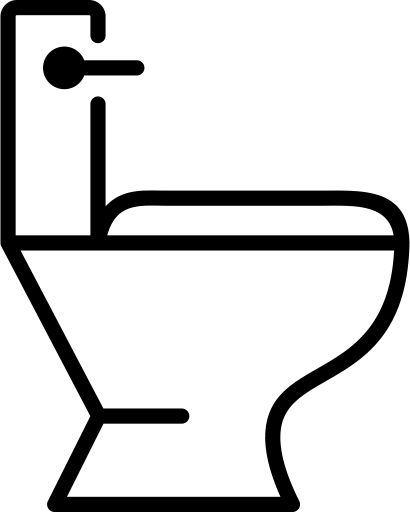}
        \\\hline\hline

  UMPNet&  0.03 & 0.07 & 0.52 & 0.02 & 0.00 & 0.19 & 0.44 & 0.00 & 0.21 & 0.08 & 0.47 & 0.00 & 0.00 & 0.04 & 0.00 & 0.06 \\
  FlowBot3D&  0.24 & 0.16 & \textbf{0.85} & 0.03 & 0.60 & 0.62 & \textbf{0.79} & 0.50 & 0.55 & 0.03 & \textbf{0.85} & 0.38 & 0.27 & 0.08 & 0.00 & 0.34 \\
  3DImplicit&  0.18 & 0.11 & 0.45 & 0.08 & 0.00 & 0.45 & 0.45 & 0.10 & 0.34 & 0.05 & 0.50 & 0.00 & 0.09 & 0.16 & 0.00 & 0.08 \\
  ManipLLM& 0.13 & 0.21 & 0.70 & 0.02 & 0.12 & 0.38 & 0.11 & 0.17 & 0.11 & 0.01 & 0.45 & 0.00 & 0.23 & 0.08 & 0.31 & 0.15\\\hline
  Ours& 0.50 &\textbf{0.47}&	0.29 &	0.34&	0.46 &	0.78&	0.47&	\textbf{0.75}&	0.40&	\textbf{0.72} &	0.63&	0.50&	0.76&	0.32& 0.31 & 0.55 \\ \hline 
  Ours(random)& 0.43 & 0.29 & 0.69 & 0.21 & 0.68 & 0.69 & 0.54 & 0.58 & 0.33 &0.45 & 0.53 &\textbf{0.75} & 0.76 &0.33 & \textbf{0.69} &0.49 \\ \hline 
  Ours (long)& \textbf{0.63} & 0.42& 0.52& \textbf{0.45}& \textbf{0.71}& \textbf{0.86}& 0.63& \textbf{0.75}& \textbf{0.66}& 0.69& 0.70& \textbf{0.75}& \textbf{0.82}& \textbf{0.34}& 0.63& \textbf{0.60}\\\hline
			\hline
    \multirow{2}{*}{\textbf{}} & \multirow{2}{*}{\textbf{}} & \multicolumn{2}{c|}{\textbf{}} & \multicolumn{12}{c}{\textbf{Test Categories}} \\

	Method
       
        &\includegraphics[height=0.035\linewidth]{images/mini_img/storageFurniture.png}
        &\includegraphics[height=0.035\linewidth]{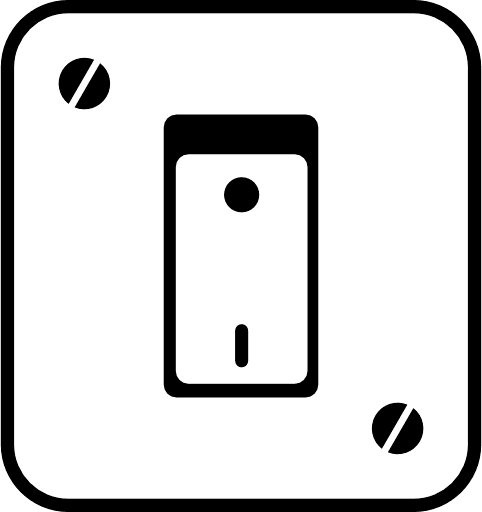}

        &\multicolumn{1}{c|}{{\textbf {AVG}} }
        &\includegraphics[height=0.035\linewidth]{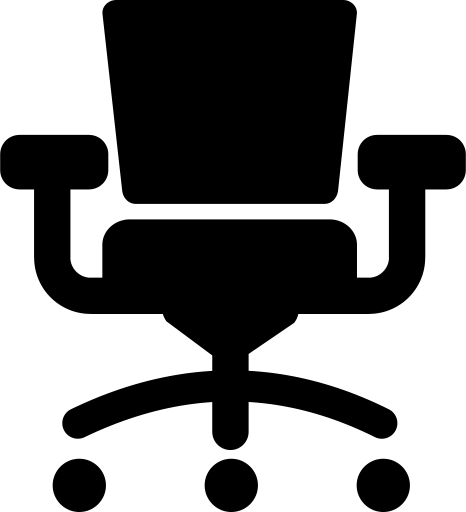}
        &\includegraphics[height=0.035\linewidth]{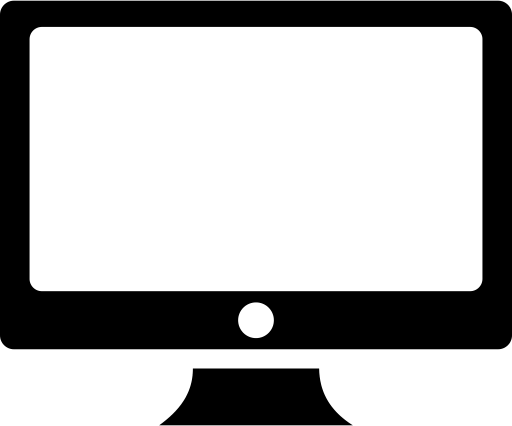}
        &\includegraphics[height=0.035\linewidth]{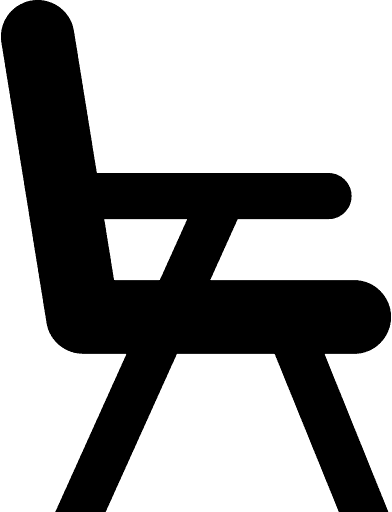}
        &\includegraphics[height=0.035\linewidth]{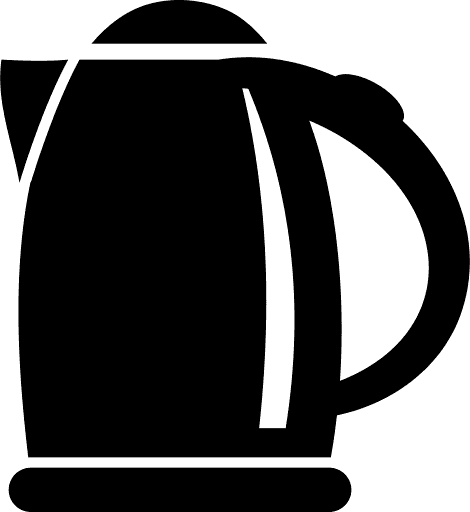}
        &\includegraphics[height=0.035\linewidth]{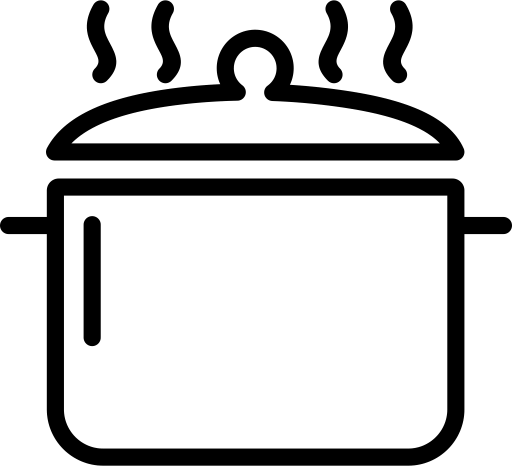}
        &\includegraphics[height=0.035\linewidth]{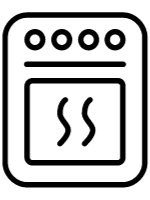}
        &\includegraphics[height=0.035\linewidth]{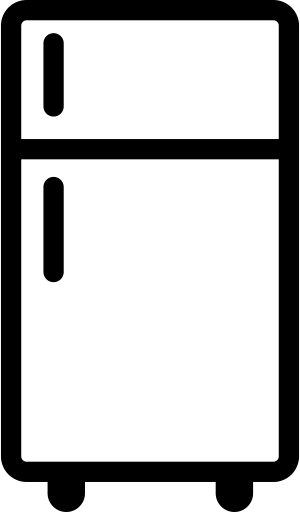}
        &\includegraphics[height=0.035\linewidth]{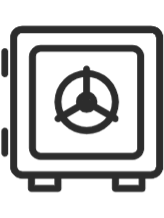}
        &\includegraphics[height=0.035\linewidth]{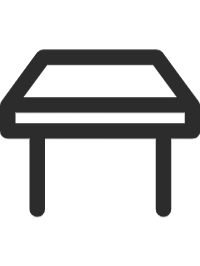}
        &\includegraphics[height=0.035\linewidth]{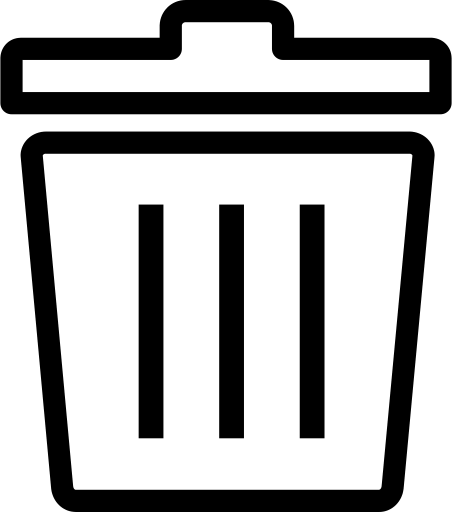}
        &\includegraphics[height=0.035\linewidth]{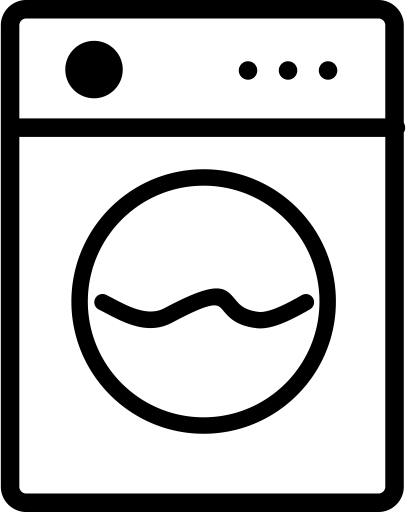}
        &\includegraphics[height=0.035\linewidth]{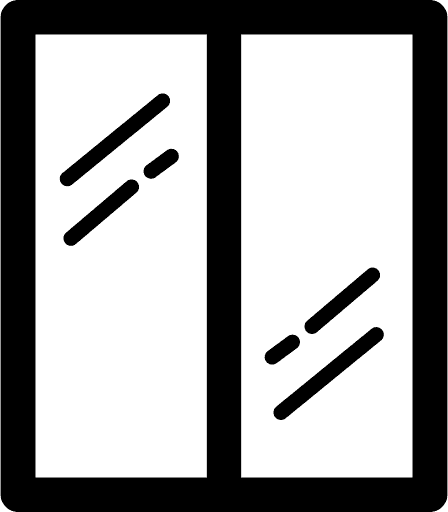}
        &{\textbf {AVG}}\\\hline\hline
  
   UMPNet &0.30 & 0.10& \multicolumn{1}{c|}{0.14}& 0.56 & 0.15 &0.07 &0.02 &0.00 &0.13 &0.50 &0.63 &0.13 &0.07 &0.33 &0.17 & 0.23\\
  FlowBot3D& 0.76 & 0.40 	&\multicolumn{1}{c|}{0.41}	&0.79 &0.17 &0.21 &0.29 &0.03 &0.13 &0.88 & \textbf{0.91} &0.53 &0.26 &0.66 &0.57 & 0.45 \\
  3DImplicit&  0.55& 0.18 & \multicolumn{1}{c|}{0.21} & 0.38& 0.17& 0.25& 0.09& 0.04& 0.18& 0.55& 0.62& 0.45& 0.10& 0.33& 0.19 &0.28\\
  ManipLLM& 0.55 & 0.25 & \multicolumn{1}{c|}{0.22} & 0.69 & 0.20 & 0.16 & 0.04& 0.03 & 0.08& 0.41 & 0.41 & 0.49& 0.19 & 0.47 & 0.50 & 0.31 \\\hline
  Ours& 0.80& 0.43&	\multicolumn{1}{c|}{0.53}& 0.88& 0.23&	0.77&	0.38	&0.11&	0.73&	0.88& 0.80& 0.35&	\textbf{0.52}& 0.64 &	0.58 &0.57\\ \hline 
  Ours(random)& 0.59 &0.29 &	\multicolumn{1}{c|}{0.52} & 0.69 &0.14 &0.53 &0.45 &0.16 &0.48 & 0.77&0.60 &\textbf{0.62}&0.46&0.52&0.36& 0.48 \\ \hline 
  Ours (long)& \textbf{0.88}& \textbf{0.63}& \multicolumn{1}{c|}{\textbf{0.65}} &\textbf{0.94}& \textbf{0.36} & \textbf{0.89} & \textbf{0.65} & \textbf{0.30} &\textbf{0.86} & \textbf{0.94} & 0.88 &0.46 &\textbf{0.52} &\textbf{0.69} &\textbf{0.61} &\textbf{0.68}\\\hline

		\end{tabular}}
	\end{center}
\end{table*}

\subsection{Simulation Experiment and Baselines} 
The affordance predictor is evaluated against image segmentation baselines, while the action proposer is evaluated in conjunction with it. Before executing actions, we measure the Union False Positive Ratio ($FPR_{Union}$) to assess the accuracy of our model's affordance masks. Manipulations are skipped and marked as failures if $FPR_{Union}$ exceeds 0.5, indicating substantial false-positive regions beyond the ground truth. The FPR is calculated as:
\begin{equation}
    \text{$FPR_{Union}$} = \frac{\text{False Positives}}{\text{Predicted Mask} \cup \text{Ground Truth Mask}}
\end{equation}

For the simulation experiments, we evaluate pull motions—such as lifting lids, opening doors, and pulling drawers—using a suction gripper in the SAPIEN environment~\cite{xiang2020sapien}, with articulated objects from PartNet-Mobility~\cite{mo2019partnet}. The experiments cover both one-step and multi-step manipulations.

\begin{itemize}
\item \textbf{One-step Manipulation}: The robot pulls 0.18 units in the direction proposed by the action proposer. Success thresholds are 0.1 units for prismatic joints and 0.1 radians for revolute joints.
\item \textbf{Multi-step Manipulation}: The robot perform seven steps, each moving 0.05 units, with impedance control applied after each step. Success thresholds are set at 0.3 units and 0.3 radians.
\end{itemize}

We further analyze our framework by testing it without post-processing but selecting a random point within the affordance mask for grasping and pulling directions. This ``random-point''  policy is evaluated using the same criteria as the one-step manipulation.

Comparison with four baselines is as follows, with all using the same simulation environment and end-effector:

\begin{itemize}
\item \textbf{UMPNet}~\cite{xu2022universal}: Predicts closed-loop action sequences for manipulating articulated objects from RGB-D images, leveraging the ``Arrow-of-Time'' concept for goal-conditioned, multi-step manipulation. It uses two components: DistNet, which predicts contact points and manipulation directions, and AoTNet, which determines whether the action advances or reverses. 

\item \textbf{Flowbot3D}~\cite{eisner2022flowbot3d}: Uses a 3D Articulation Flow (3DAF) vector field to predict point-wise motion directions. The ArtFlowNet model selects the interaction point based on the largest flow magnitude and the corresponding direction for end-effector positioning.

\item \textbf{Implicit3D}~\cite{zhong20233d}: Extends the 2D Transporter model to 3D, discovering temporally consistent keypoints in point cloud sequences. We use its 3D keypoints to determine the end-effector's pose.

\item \textbf{ManipLLM}~\cite{li2024manipllm}: A multi-modal system using a language prompt and RGB image to predict the end-effector's pose through a structured chain-of-thought process. Once initial contact is established, subsequent waypoints are dynamically planned using active impedance adaptation.
\end{itemize}

\subsection{Quantitative Evaluation in Simulator}

Following the settings from~\cite{li2024manipllm}, the baseline evaluation focuses on the initial movement phase, which is crucial for assessing a model's ability to understand and respond to an object's initial state—key for successful long distance manipulation. All models are trained on a standardized dataset of 5 objects per category across 18 categories and evaluated under the same conditions. For ManipLLM, we use its pretrained model as it is trained on a larger set of objects and affordance data than ours. As a result, the train and test categories in Table \ref{tab:pull} are not directly comparable, and the results reflect ManipLLM’s generalization capability rather than performance on our dataset.

The results in Table \ref{tab:pull} reveal that our approach consistently achieves high success rates, averaging 53\% on seen and 57\% on unseen object categories. While this may seem counterintuitive, the higher success rate on unseen categories can be attributed to structural similarities. Many unseen objects share articulation mechanisms (e.g., doors, lids) with those in the training set, despite belonging to different categories. Variation within training categories may also make seen objects harder to manipulate than structurally similar yet unseen ones.

The aforementioned random-point policy achieve success rate of 52\% on seen and 48\% on unseen categories. However, this approach is less effective for real-world applications, as points on protruding features (e.g. handles) or near revolving joints (e.g. door hinges) often lead to poor attachment or excessive force during manipulation, reducing success rates for certain categories.

Multi-step manipulations demonstrate the highest success rates, achieving 65\% on train and 68\% on test categories. This underscores the value of adaptive control in handling a variety of articulated objects, highlighting the robustness and adaptability of our algorithm. Multi-step evaluations are not conducted for baselines as they already showed lower single-step success rates. Given that single-step failure rates are significant, applying these baselines to multi-step tasks would not provide meaningful insights.

\begin{figure}[t]
    \begin{center}
        \includegraphics[width=1.0\linewidth]{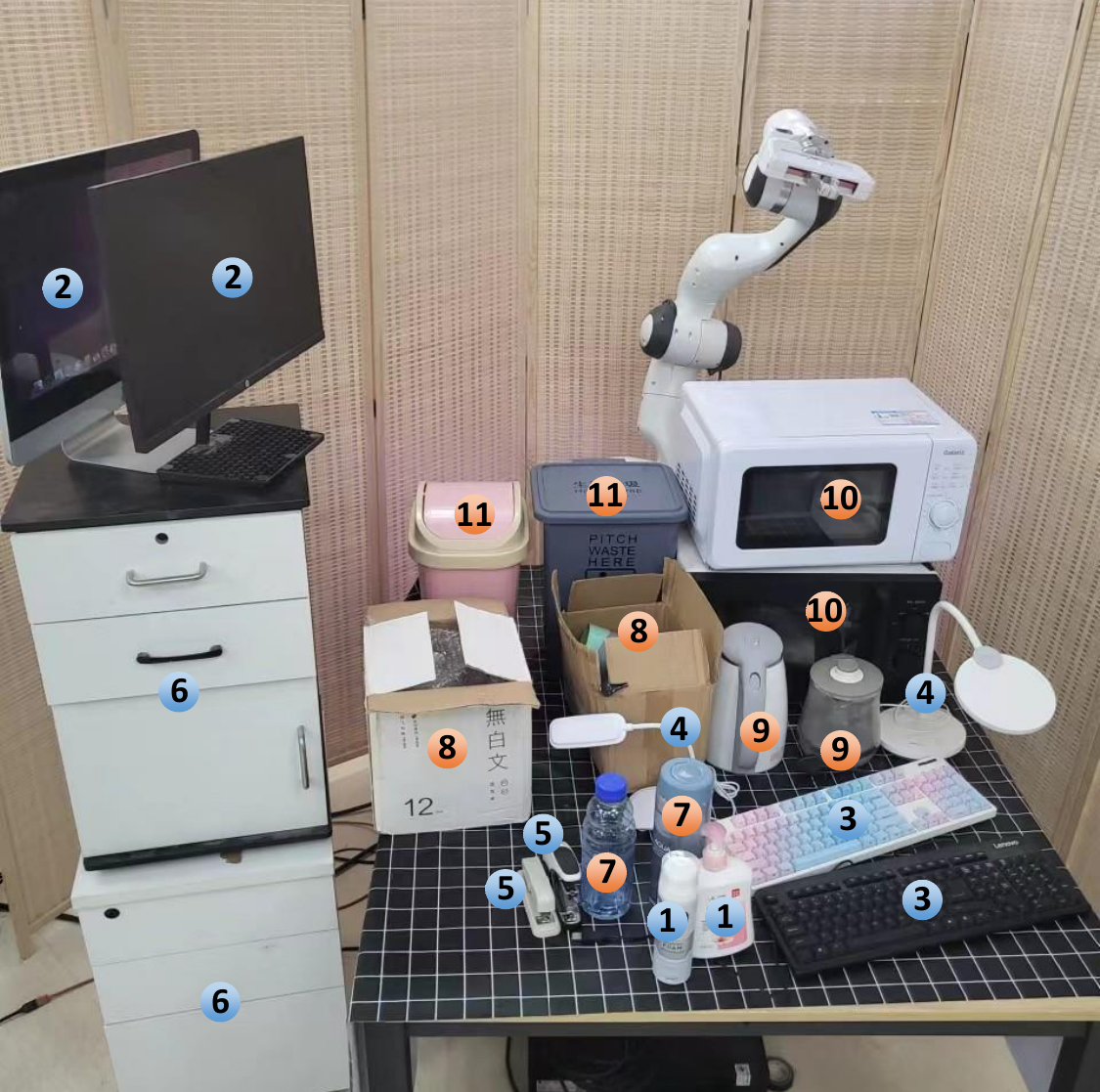}
    \end{center}
    \caption{We test 11 objects in our real-world experiment, numbered 1–11 as listed in Table \ref{tab:pushing_and_pulling}. Objects 1–6 belong to the pushing category, while 7–11 are in the pushing category.}
    \label{fig:real-world-exp}
\end{figure}

\begin{table}[h]
    \centering
    \small
    \caption{Real world experiments for pushing and pulling categories.}
    \label{tab:pushing_and_pulling}
    \begin{tabular}{|c|*{6}{p{0.5cm}|}}
    \hline
    \multicolumn{7}{|c|}{Pushing Categories} \\
    \hline
    Object Category & \multicolumn{1}{c|}{\includegraphics[height=0.6cm]{images/mini_img/dispenser.png}} &
    \multicolumn{1}{c|}{\includegraphics[height=0.6cm]{images/mini_img/display.png}} &
    \multicolumn{1}{c|}{\includegraphics[height=0.3cm]{images/mini_img/keyboard.png}} &
    \multicolumn{1}{c|}{\includegraphics[height=0.6cm]{images/mini_img/lamp.png}} &
    \multicolumn{1}{c|}{\includegraphics[height=0.6cm]{images/mini_img/stapler.png}} &
    \multicolumn{1}{c|}{\includegraphics[height=0.6cm]{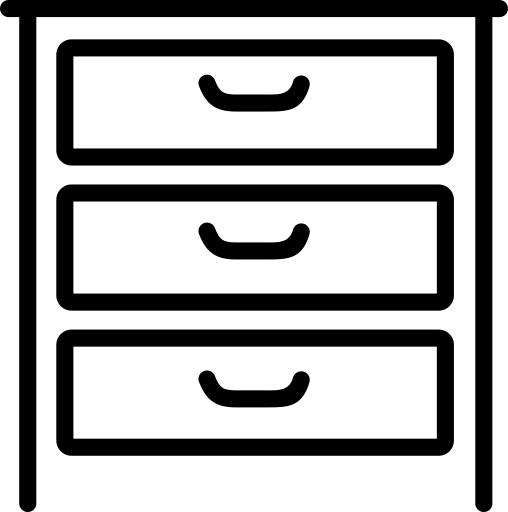}}\\
    \hline
    Success/Total & 5/6 & 3/6 & 6/6 & 5/6 & 4/6 & 5/6\\
    \hline
    Distance(m) & 0.01 & 0.11 & 0.01 & 0.12 & 0.02 & 0.18 \\
    \hline
    \multicolumn{7}{|c|}{Pulling Categories} \\
    \hline
    Object Category & \multicolumn{1}{c|}{\includegraphics[height=0.6cm]{images/mini_img/bottle.png}} &
    \multicolumn{1}{c|}{\includegraphics[height=0.6cm]{images/mini_img/box.png}} &
    \multicolumn{1}{c|}{\includegraphics[height=0.6cm]{images/mini_img/kettle.png}} &
    \multicolumn{1}{c|}{\includegraphics[height=0.5cm]{images/mini_img/microwave.png}} &
    \multicolumn{1}{c|}{\includegraphics[height=0.6cm]{images/mini_img/trashcan.png}} &
    \textbf{Avg} \\
    \hline
    Success/Total & 5/6 & 6/6 & 6/6 & 5/6 & 5/6 & \textbf{5/6} \\
    \hline
    Distance(m) & 0.23 & 0.09 & 0.15 & 0.16 & 0.18 & \textbf{0.11} \\
    \hline
    \end{tabular}
\end{table}

\begin{figure*}[t]
    \centering
    \includegraphics[width=\textwidth]{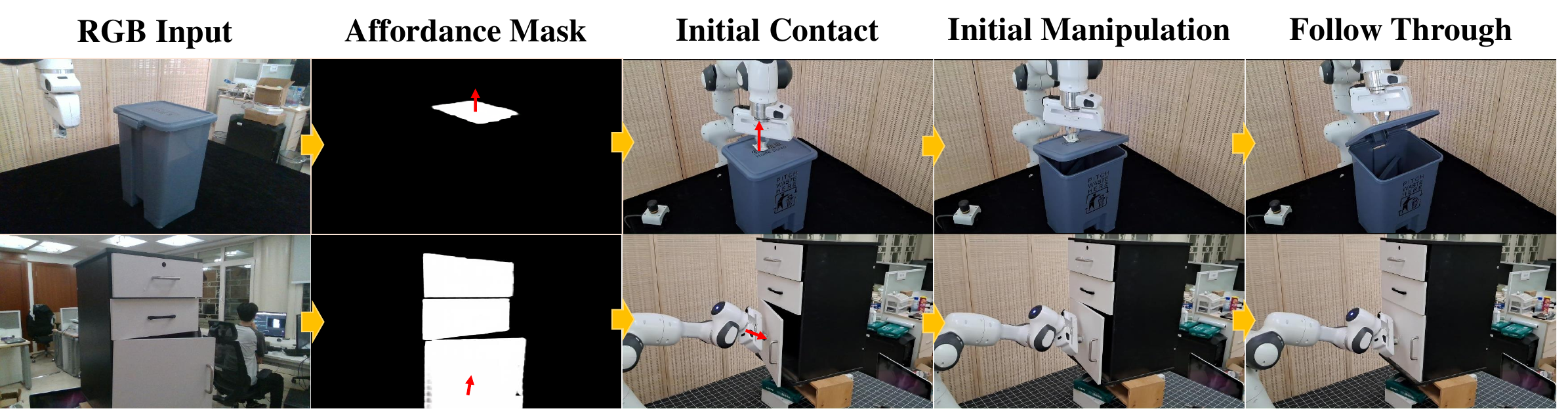}
    \caption{This figure demonstrates the real object experiment process. Above and below show pulling and pushing manipulation, respectively.}
    \label{fig:real-object-confidence}
\end{figure*}

\subsection{Quantitative Evaluation in Real-world}

We evaluate our framework using a real robot interacting with physical objects. The experiment covers five pulling and six pushing object categories, with two object instances per category. Each object is tested three times, totaling six trials per category. RGB and depth images are captured using a RealSense D415, and our model predicts affordance masks, which are projected into 3D space to determine the end-effector’s target pose.

We use a Franka Emika Panda robot, originally equipped with a parallel gripper. To replicate suction-based interactions from simulation, we convert it into a sticky gripper by taping its closed fingers with double-sided foam. Although the grip isn't strong enough to hold heavy objects long-term, it is sufficient to demonstrate manipulation performance.

The results of real-world experiment are provided in Table~\ref{tab:pushing_and_pulling}. We observe that our method shows no significant sim-to-real gap thanks to the synthetic image dataset. The model achieves a high success rate, excelling particularly in manipulating objects with well-defined boundaries and low actuation force requirements, such as boxes or keyboards. However, it occasionally struggles with objects featuring transparent or small targets like dispensers, or those with ambiguous boundaries like staplers. Additionally, accurate affordance segmentation for monitors can be challenging due to light reflections on the screen.

Using real image prompts for real-world testing, as opposed to simulated prompts in simulation, our results in Figure~\ref{fig:real-object-confidence} show that the model generalizes well to real-world robotic scenarios.


\begin{table}[h!]
  \begin{center}
  \caption{Ablation Study. All models are tested on real images only.}
    \label{tab:ablation_compasion}
    \begin{tabular}{l|c|c|c} 
    \textbf{Model} & \textbf{SegGPT} & \textbf{Ours(sim)} & \textbf{Ours} \\
      \hline
      \textbf{F}\textsubscript{1}\textbf{score}
      & 45.1 & 49.2 & \textbf{80.3}\\
      \textbf{mIoU}
      & 36.2 & 40.2 & \textbf{73.4}\\
    \end{tabular}
  \end{center}
\end{table}

\subsection{Ablation Study}
Our ablation study investigates the impact of including real images in the training set. We compare three configurations: SegGPT, ours trained only on simulated images, and ours trained with both real and simulated images.

Table \ref{tab:ablation_compasion} shows that even a small number of real images significantly enhance SegGPT's generalization ability for affordance segmentation. Specifically, including real images in the training set lead to a 31.1\% increase in the F\textsubscript{1} score and a 33.2\% increase in the mIoU value. Given the importance of real-world images for manipulation tasks, our results strongly support the effectiveness of incorporating real data into the training process for real-world applications.

\subsection{Limitations and failure cases}

Our framework occasionally fails to produce affordance masks, particularly for small objects or novel viewpoints, likely due to limited training data or ambiguity between the prompt and query images. Expanding the dataset or adopting a few-shot learning approach with multiple prompts could mitigate this. Additionally, the reliance on one-shot prompts limits effectiveness for unseen categories when a suitable prompt is unavailable; integrating a similarity-based prompt retrieval system could address this issue.

Challenges also arise with transparent objects, such as kitchen pot lids, where transparency hinders both segmentation and normal map estimation, leading to lower success rates (e.g., Kitchen Pot category in Table \ref{tab:pull}). Training on video data to leverage temporal consistency could improve segmentation robustness.

Finally, selecting optimal manipulation points remains challenging. Our approach targets the center of the mask after filtering, but this may not be ideal for heavy objects or revolute joints. Incorporating more detailed prompts with annotated manipulation points could improve precision.

\section{Conclusion}
We present a vision-driven approach to robotic manipulation that integrates in-context affordance prediction with responsive control. By fine-tuning a large vision model and leveraging normal vectors for interaction guidance, our system efficiently identifies manipulatable regions using a single RGB image and prompt pair, eliminating the need for robot data. Experiments prove robust affordance prediction, enabling precise contact localization and motion direction estimation.

While our experiments showcase suction-based manipulation, it is adaptable to other grippers. Once the affordance predictor localizes the actionable region, one can replace the suction-oriented action proposer into a 6D grasp-pose estimator (e.g., GraspNet-1b~\cite{fang2020graspnet}) for parallel-jaw grasping on the segmented region. The impedance control stage could remain relevant for safe, compliant manipulation.

\section{Acknowledgments}
We thank Dr. Xinlong Wang and Dr. Xiaosong Zhang for  their technical suggestions that improved our training design. This project was supported by the National Youth Talent Support Program (8200800081), National Natural Science Foundation of China (62376006) and National Natural Science Foundation of China (62136001)

\bibliographystyle{IEEEtran}
\bibliography{IEEEabrv, reference}

\end{document}